\documentclass[conference]{IEEEtran}
\usepackage{multirow}
\usepackage{amsmath,algcompatible}
\usepackage{algorithm}
\usepackage{mathtools}
\usepackage{adjustbox}
\usepackage{breqn}
\usepackage[T1]{fontenc}

\DeclareMathOperator*{\argmin}{argmin}
\algnewcommand\INPUT{\item[\textbf{Input:}]}%
\algnewcommand\OUTPUT{\item[\textbf{Output:}]}%

\usepackage{booktabs}
\usepackage{cite}
\usepackage{amssymb,amsfonts}
\usepackage{graphicx}
\usepackage{textcomp}
\usepackage{xcolor}
\def\BibTeX{{\rm B\kern-.05em{\sc i\kern-.025em b}\kern-.08em
    T\kern-.1667em\lower.7ex\hbox{E}\kern-.125emX}}

\begin{document}

\makeatletter
\newcommand{\linebreakand}{%
  \end{@IEEEauthorhalign}
  \hfill\mbox{}\par
  \mbox{}\hfill\begin{@IEEEauthorhalign}
}
\makeatother

% \title{A Comprehensive Demand Forecasting Framework for Ecommerce Advertising Products}
\title{A Comprehensive Forecasting Framework based on Multi-Stage Hierarchical Forecasting Reconciliation
and Adjustment
}

\author{
    \IEEEauthorblockN{Zhengchao Yang}
    \IEEEauthorblockA{
    \textit{Walmart Global Tech}\\
    Sunnyvale, California, USA \\
    zhengchao.yang@walmart.com}
    \and
    \IEEEauthorblockN{Mithun Ghosh}
    \IEEEauthorblockA{
    \textit{Walmart Global Tech}\\
    Sunnyvale, California, USA \\
    mithun.ghosh@walmart.com}
    \and
    \IEEEauthorblockN{Anish Saha}
    \IEEEauthorblockA{
    \textit{Walmart Global Tech}\\
    Sunnyvale, California, USA \\
    anish.saha@walmart.com}
    % \and
    \linebreakand % <------------- \and with a line-break
    \IEEEauthorblockN{Dong Xu}
    \IEEEauthorblockA{
    \textit{Walmart Global Tech}\\
    Sunnyvale, California, USA \\
    dong.xu@walmart.com}
    \and
    \IEEEauthorblockN{Konstantin Shmakov}
    \IEEEauthorblockA{
    \textit{Walmart Global Tech}\\
    Sunnyvale, California, USA \\
    konstantin.shmakov@walmart.com}
    \and
    \IEEEauthorblockN{Kuang-chih Lee}
    \IEEEauthorblockA{
    \textit{Walmart Global Tech}\\
    Sunnyvale, California, USA \\
    kuangchih.lee@walmart.com}
}

\maketitle


\begin{abstract}
Ads demand forecasting for Walmart Connect’s ad products plays a critical role in enabling effective resource planning, allocation, and overall management of ads performance. In this paper, we introduce a comprehensive demand forecasting system that tackles the hierarchical time series forecasting problem in business settings. 
Though traditional hierarchical reconciliation methods ensure forecasting coherence, they often trade off accuracy for coherence especially at lower levels and fail to capture the seasonality patterns unique to each time-series in the hierarchy. Thus, we propose a novel framework “Multi-Stage Hierarchical Forecasting Reconciliation and Adjustment (Multi-Stage HiFoReAd)” to address the challenges of preserving seasonality, ensuring coherence, and improving forecasting accuracy. Our system first utilizes diverse modeling techniques, ensembled through Bayesian Optimization (BO), achieving individual base forecasts. The generated base forecasts are then passed into the Multi-Stage HiFoReAd framework. The initial stage refines the hierarchy using Top-Down forecasts and “harmonic alignment.” The second stage aligns the higher levels’ forecasts using MinTrace algorithm, following which the last two levels undergo “harmonic alignment” and “stratified scaling”, to eventually achieve accurate and coherent forecasts across the whole hierarchy. Our experiments on Walmart's internal Ads-demand dataset and 3 other public datasets, each with 4 hierarchical levels, demonstrate that the average Absolute Percentage Error (APE) from the cross-validation sets improve from 3\% to 40\% accross levels against BO-ensemble of models (LGBM, MSTL+ETS, Prophet) as well as from 1.2\% to 92.9\% against State-Of-The-Art models. In addition, the forecasts between all hierarchical levels are proved to be coherent. The proposed framework has been deployed and leveraged by Walmart's ads, sales and operations teams to track future demands, make informed decisions and plan resources strategically. 

% Ads demand forecasting for Walmart Connect’s ad products plays a critical role in enabling effective resource planning, allocation, and overall management of ads performance. 
% In this paper, we introduce a comprehensive demand forecasting system that tackles the hierarchical time series forecasting problem in business settings. 
% Though traditional hierarchical reconciliation methods ensure forecasting coherence, they often tradeoff accuracy for coherence, and fail to capture the seasonality patterns unique to each time-series in the hierarchy. 
% Thus, we propose a novel framework “Multi-Stage Hierarchical Forecasting Reconciliation with Adjustment (Multi-Stage HiFoReAd)” to address this challenges by preserving seasonality, ensuring coherence, and improving forecasting accuracy.  
% Utilizing forecasts ensembled through Bayesian Optimization (BO),
% the Multi-Stage HiFoReAd framework refines the hierarchical forecasts through Top-Down and 
% MinTrace reconciliation, coupled with our “harmonic alignment” and “stratified scaling”, to achieve accurate and coherent forecasts. 
% Our experiments on real-world data containing 12,000 time series in 4 hierarchical levels demonstrated that the average Absolute Percentage Error (APE) improved by 16-40\% (depending on the hierarchy) against BO-ensemble of classical models, and 25-50\% over SOTA transformer based models.
% This proposed methodology framework has been evaluated and leveraged by Walmart Connect sales and operations team to track future demands, make informed decisions and plan resources strategically. 
\end{abstract}




\section{Introduction}
Ads demand forecasting involves the process of predicting the future demands for advertising products across various channels and platforms. Today's hyper-competitive business landscape makes accurate demand forecasting important for any organization’s success \cite{yan2016manufacturer}. 
This is instrumental in devising long-term and short-term strategies to evaluate and improve a company's performance \cite{zhang2020request}.
Moreover, the original consumer's demand is translated into an order from the retailer to replenish its stock, which in-turn necessitates an order on the next stage (e.g. manufacturer), and so on until the end of the chain is reached.
There can be several amplifications of demand variance through the stages, termed the “Bullwhip Effect” \cite{lee1997bullwhip}, making it crucial to have accurate demand forecasts. 

Walmart's ads demand forecasting requires the estimation of the future sales for each business segment and demands of ad products within a specified period of time, to optimize the strategies and ensure effective allocation of resources to meet consumer engagement goals. Oftentimes, these forecasts need to be generated within a hierarchical structure, which could represent product families, geographic regions, or/and business segments, and so on \cite{abolghasemi2022model, hyndman2014optimally}. The merit of effective hierarchical forecasting resides in its capacity to furnish decision-support information to stakeholders within distinct organizational functions and managerial levels, consequently helping Walmart pilot its execution plans for reaching desired YoY growth targets, and ultimately boosting the company’s bottom operational capability.

% It has become evident that m
Most existing methods focus on conducting forecasting tasks at all hierarchical levels independently, which fail to capture the interdependence and shared influences within a hierarchical structure during the forecasting process and therefore lead to incoherent forecasts. 
Also, dealing with forecasting of large-scale hierarchical time series remains a formidable challenge, even with the art and science of existing forecasting methods. 
The interplay of factors such as: 
(1) gauging market trends and seasonality, 
(2) handling data noise and quality, time series randomness, 
(3) forecast incoherence throughout the hierarchy, 
(4) model selection, hyper-parameter tuning etc., 
(5) requirement of substantial computational resources, 
makes it increasingly complex to make scalable, accurate, and coherent forecasts across the hierarchy. 
Existing forecasting methods \cite{taylor2018forecasting, wang2020short, yenidougan2018bitcoin, zhang2020request}  often fall short, only focusing on one or a few aspects of these challenges, which could lead to untimely and inaccurate forecasts, and consequently, substantial financial losses.

These challenges motivate us to propose a comprehensive framework for in-house Ads spend demand forecasting, with our main contributions being the following: 
\begin{itemize}
\item We propose a novel framework “Multi-Stage Hierarchical Forecasting Reconciliation and Adjustment (Multi-Stage HiFoReAd)” to obtain coherent forecasts throughout the hierarchy, while preserving the intrinsic trend and seasonality of each time-series through “harmonic alignment” and “stratified scaling”, and maintaining accuracy. 
\item We provide an optimization framework to obtain an ensemble of State-Of-The-Art ML forecasting models, to capture a wide range of time series patterns and uncertainty to provide robust forecasts.
% while incorporating domain-specific knowledge.
\item We bridge the gap between scalability, robustness, and accuracy by leveraging distributed computing to parallelize the modeling process. 
\item We demonstrate our framework's capability to provide coherent and accurate forecasts on 4 real-world datasets, and compare its improvements over State-of-the-Art forecasting and reconciliation models. 
\end{itemize}

We delve into each component, outlining the data preprocessing/segmentation and distributional modeling in sec. \ref{sectionOverview}, \ref{sectionPrelim}, the Bayesian Optimization based ensemble and Hierarchical Forecasting Reconciliation in sec. \ref{sectionMSHFRA}. 
We also discuss the implementation of this framework in practical scenarios, sharing insights from our case study that demonstrates the framework’s efficacy in sec. \ref{sectionExperiment}.

\section{Literature Review}
\label{litReview}
% In the fast-paced domain of advertising, demand forecasting is a critical component for all the planning teams in any company \cite{yan2016manufacturer}. 
% This information is instrumental in devising both long-term and short-term strategies to evaluate and improve a company's performance \cite{zhang2020request}.
% Moreover, the original consumer's demand is translated into an order from the retailer to replenish its stock, which in-turn necessitates an order on the next stage (e.g. manufacturer, etc.), and so on until the end of the chain is reached.
% There can be several amplifications of demand variance through the stages, termed the ``Bullwhip Effect'' \cite{lee1997bullwhip}, making it crucial to have accurate demand forecasts. 
Forecasting can be computationally expensive, particularly in the AdTech industry, where both the number and the length of time series can grow significantly over time. Various studies \cite{karmy2019hierarchical, pennings2017integrated, villegas2018supply} underscore the inadequacy of independent level's forecasts in capturing the intricate relationships within organizational and business hierarchies.
Additionally, many time series in this domain contain substantial amounts of intermittent data, making traditional models less applicable and studies from large-scale forecasting competitions suggest that combining models instead of selecting one ``best'' model can improve accuracy and robustness \cite{clemen1989combining, makridakis1982accuracy, makridakis1993m2}.

Thus, we first incorporate an Bayesian Optimization based ensemble of the best-performing ML models from various domains, to achieve forecasts as robust and accurate as possible across the whole hierarchical structure. 

Additionally, in hierarchical structures, independently forecasting each time series has been a common practice in industries. These models typically predict outcomes at top or lower levels without explicitly considering dependencies that exist between different levels. Once the forecasts are made, Top-Down (TD), Bottom-Up (BU) or Middle-Out \cite{abolghasemi2022model, babai2022demand, li2018greedy, rostami2015non, zotteri2007model} methods are applied to dis-aggregate the top levels' forecasts to generate the bottom levels', or to generate top levels' forecasts by aggregating the bottom levels'. 
Though these methods guarantee the coherence between levels of the hierarchy, they often fail to incorporate data from all levels, resulting in the loss of valuable information and degradation of prediction accuracy in many real-word scenarios.

Classic reconciliation methods \cite{ben2019regularized, wickramasuriya2018optimal}, such as MinTrace, ERM, etc., aim to address this challenge by adjusting the individual forecasts at multiple levels simultaneously to ensure coherence. 
However, these techniques have some notable shortcomings: (1) assumption for unbiased forecasts and Gaussian noises, which often do not hold in real-world scenarios with complex relationships, (2) struggle to handle intermittence or fluctuations in the bottom level time series, eventually leading to sub-optimal reconciled forecasts, (3) computational complexity of certain reconciliation algorithms and their challenges in large-scale applications. 
Other machine learning based reconciliation methods \cite{mancuso2021machine, spiliotis2021hierarchical} are often evaluated on small-scale volumes of time series, due to their limited scalability while estimating the relationship between levels. Specifically their inability in handling large scale variance/covariance matrices for the hierarchy, whose dimensionality explodes as the number of time series increases, hinders the adoption of machine learning based methods in industry. 

Our proposed framework, dubbed ``HiFoReAd'', addresses these limitations by decomposing the reconciliation task into four main stages to handle scalability, preserve intrinsic uniqueness and patterns of individual time series, and maintain coherence across the hierarchy. 

\section{Methodological Framework}

\subsection{Methodology Overview}
\label{sectionOverview}
\begin{figure}[h]
  \centering
  \includegraphics[width=\linewidth]{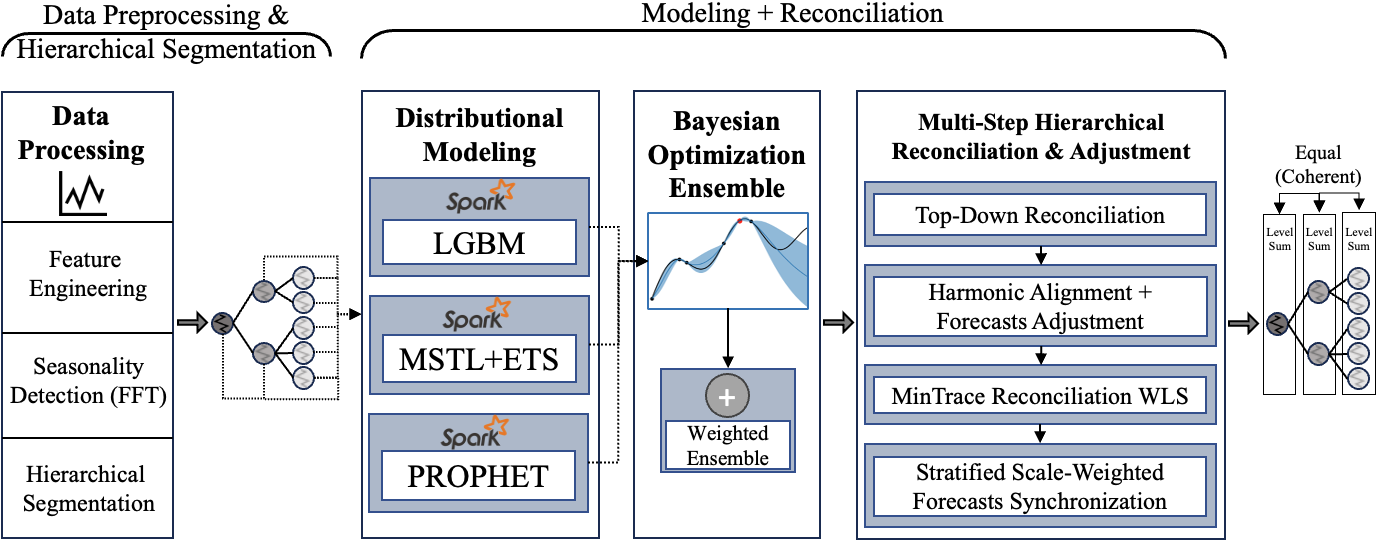}
  \caption{The Framework Design of the Proposed Demand Forecasting System}
  \label{fig:framework}
\end{figure}
In Figure \ref{fig:framework}, we provide an overview of our framework which performs preprocessing, distributional modeling and Bayesian Optimization ensemble, followed by the Hierarchical Forecasting Reconciliation step to produce coherent and accurate multi-horizon forecasts for all hierarchical time series.
% In the following sections, we introduce how the design contributes to accurate and effective forecasting. 

\textbf{Data Preprocessing \& Hierarchical Segmentation} is the foundational step in our framework. 
This incorporates important temporal features (sec. \ref{sectionPrelim}) and domain-knowledge to divide the historical time series data into a hierarchy (Figure \ref{fig:hierarchy}), which is utilized in the Hierarchical Forecasting Reconciliation stage to enhance forecasting performance. 
This ensures optimal forecasting outcome by training specific forecasting models on relevant sets of data. 

\textbf{Distributional Modeling} component leverages Apache Spark, a distributed computing system, to parallelize the training of time series models (Prophet, MSTL, LGBM) as within GCP clusters to speed up the training process.

\textbf{Bayesian Optimization-base Ensemble} plays a pivotal role in our framework by enhancing the robustness and accuracy of forecasts. 
We incorporate Bayesian Optimization to automatically optimize the ensemble weights of individual models so as to leverage the strength of each constituent model for better forecasting results (sec. \ref{sectionPrelim}).

\textbf{Multi-Stage Hierarchical Forecasting Reconciliation and Adjustment} is the most critical module in our framework. 
It allows us to take the robust and accurate forecasts from the BO ensemble procedure at each level of the hierarchy, and perform reconciliation across different levels to ensure coherence (each level of the tree adds up to same value), while also further improving the forecasting accuracy (sec. \ref{sectionMSHFRA}). 
% The details are further elaborated in the following section. 

This design's objective is to strike a balance between business targets and precision of forecasting methods, while utilizing large scale data within industry settings. 
% In the subsequent sections, we delve deeper into each of these components, discussing methodologies, implementation details, and case studies that highlight the effectiveness of our comprehensive time series forecasting framework.

\subsection{Preliminaries}
\label{sectionPrelim}
The challenge for hierarchical forecasting lies in handling granular time series at lower levels while ensuring coherence and accuracy in forecasting at higher hierarchical levels. 
Suppose we have $N$ granular time series, denoted as $\textbf{y}_i(t|0 \leq t \leq T)$, where $i$ ranges from 1 through $N$ representing individual time series, and $t$ denotes the time step. 
The granular time series are segmented and aggregated into their parent levels, forming a hierarchical structure denoted as $L_1, L_2, \ldots, L_{K}$, where $K$ is the total number of levels, with the most granular time series in level $L_K$. 
This hierarchical structure is unique to each dataset. An example for Walmart's ads spend data is show in Figure \ref{fig:hierarchy} - constructed based on traffic volume, business sects, product types and individual products. 

\begin{figure}[h]
  \centering
  \includegraphics[width=\linewidth]{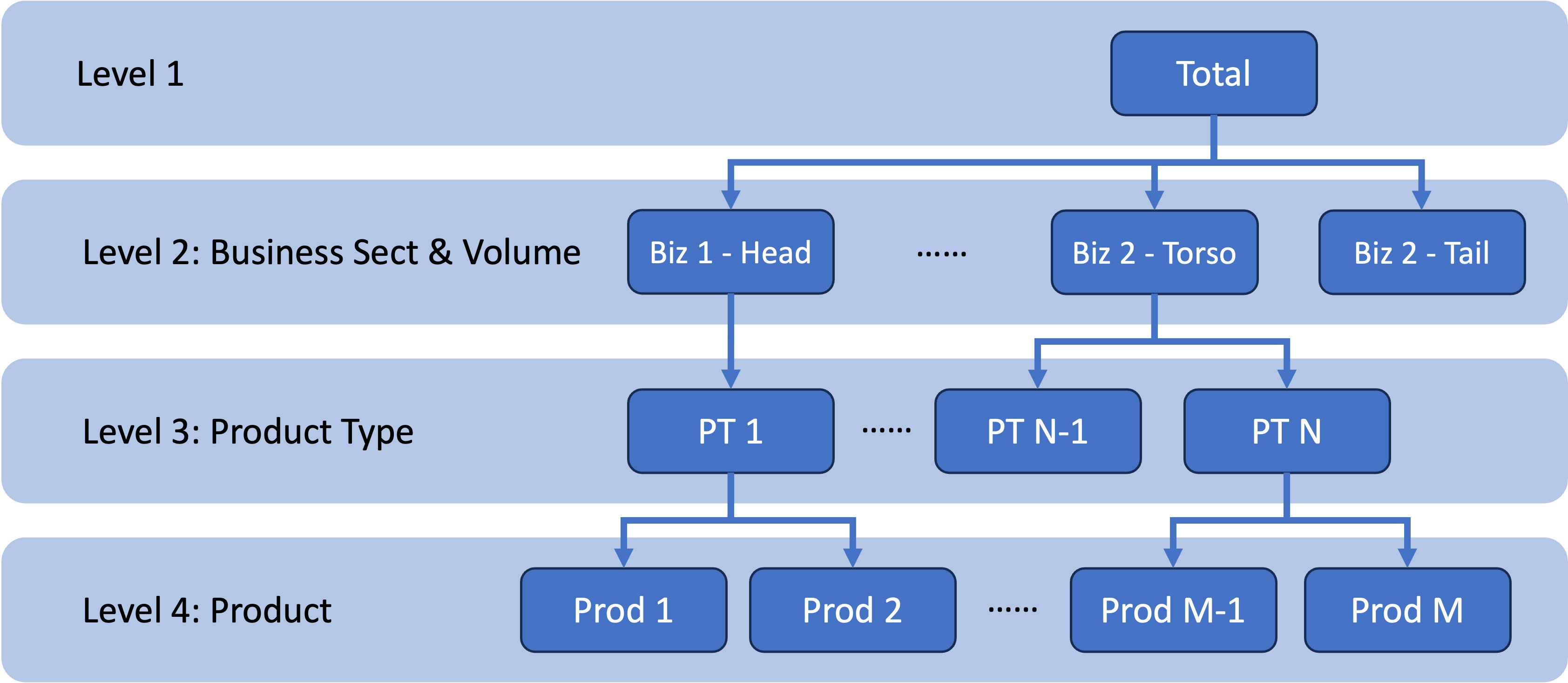}
  \caption{The Hierarchical Time Series Structure with 4 Levels}
   \label{fig:hierarchy}%
\end{figure}

At each level $L_k$, $1 \leq k \leq K$, the goal is to generate accurate forecasts, denoted as $\tilde{\textbf{y}}^{k}_i(T+1:T+h|0:T)$, for each time series $i$, $1\leq i \leq N_{L_k}$, $h$ time steps ahead, given the historical information up to time $T$. 
In addition, the coherence is guaranteed, such that:
% $\tilde{\textbf{y}}^{k}_i(T+1:T+h|0:T) = \sum^{J}_{j=1}\tilde{\textbf{y}}^{k+1}_j(T+1:T+h|0:T)$, where $j \in \textbf{Children}(i)=J$.
$$
\tilde{\textbf{y}}^{k}_i(T+1:T+h|0:T) = \sum^{J}_{j=1}\tilde{\textbf{y}}^{k+1}_j(T+1:T+h|0:T)
$$
$$
\text{where, } j \in \textbf{Children}(i)=J
$$

To improve accuracy and robustness, we find an optimal combination of the following foundational models: \verb|LGBM| \cite{ke2017lightgbm}, \verb|MSTL|\cite{bandara2021mstl}, and \verb|Prophet| \cite{taylor2018forecasting},
taking advantage of the variability in performance of individual models.
We provide explicit temporal, lag and \textit{seasonality} features to these models by computing a periodogram using Fast Fourier Transform and selecting the peaks in spectral density. 
% We also provide various temporal and lag features, engineered for these models.

% The critical nature of problem makes accurate time series forecasting essential. Since the performance of a single predictor can be highly variable due to shifts in the underlying data distribution, we aim to benefit from the diversity of multiple well-established forecasting methods. 
% To that end, we leverage multiple predictive models, to increase accuracy and robustness, as shown in Table~\ref{tab:baseModels}. 

% To model a time-series $X=\{x_1, x_2, ..., x_T\}$, where $x_t$ is a vector of $m$ input features observed at time $t$, we train each foundation model to make a $h$ step forecast into the future as $y_{T+1:T+h}$:
% $$\hat{y}(T+1:T+h) = f_{h}(x_{T-w},...,x_{T}, y_{T-w},...,y_{T},)$$
% where $\hat{y}$ is the target variable, $h$ is the prediction horizon, and $f_{h}$ is the learnt model.

\begin{figure}
  \centering
  \includegraphics[width=\linewidth]{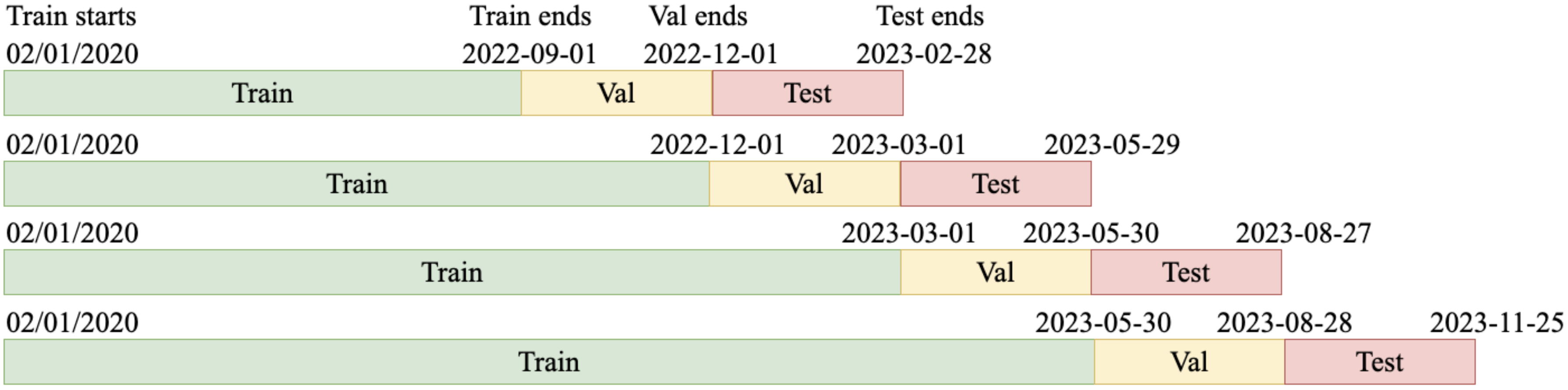}
  \caption{Back-testing framework for Walmart Connect Ads Demand data, consisting of 4 cross-validation periods containing test and validation sets.}
  \label{fig:backtest}
\end{figure}

Train-val-test splits and cross validation through \textit{back-testing} are done to bootstrap the training set and estimate the test error.
All datasets are split into 4 CV sets, with the val and test periods having a horizon $h$, as shown in Figure~\ref{fig:backtest}.

The optimal hyper-parameters ${\phi^{i}_M}^{*}$, for each model $M$ and individual time series $i$, are then chosen by minimizing the average APE over the four cross-validation sets as follows:
$${\phi^{i}_M}^{*} = \argmin_{\phi_M} \sum_{v=1}^4{\frac{1}{4} \cdot \text{APE}_M(\textbf{y}_i^{(v)}, \tilde{\textbf{y}}_i^{(v)})} $$
where $\phi_M$ is model $M$'s hyper-parameters, and $\text{APE}_M$ is its absolute percentage error (sec. \ref{sectionExperiment})
% , 
% $\text{APE}_M = \frac{1}{h}\sum_{t=1}^h\frac{|\sum_{t}{A_t}-\sum_{t}{F_t}|}{\sum_{t}{A_t}}$
% , 
for each validation-set $v$.
% Here we calculate the weighted average of $\text{MAPE}$ with weights $w_i$ for each validation period, such that $\sum_{i=1}^4{w_i}=1$.

% With this framework in place, we obtain the optimal performance for making univariate forecasts for the time-series with each foundation model.

\subsection{Multi-Stage Hierarchical Forecasting Reconciliation and Adjustment}
\label{sectionMSHFRA}
Within our framework, we integrate 3 machine learning models for each time series at each level in the hierarchy. Rather than simple average of models (which dilutes the impact of a superior model), we enhance predictive performance by the utilization of a weighted ensemble. This not only mitigates over-fitting but also imparts robustness against outliers. Additionally, the weights assigned to each model help explain their importance in shaping the final prediction.

To determining the optimal weight for each model we use Bayesian Optimization. Treating the cost function as a black box, our framework seeks to optimize the objective function denoted as $g(\textbf{x})$, where $\textbf{x}$ corresponds to the weight vector. This weight optimization is expressed as:
$$\textbf{x}^* = \mathrm{arg max}_{\textbf{x}\in \textbf{X}}g(\textbf{x})$$
The uncharted objective function is approximated through a surrogate model, commonly the Gaussian process (GP), which establishes a prior $p(g)$ delineating the objective function. 
The GP not only provides an initial estimation of $g(\textbf{x})$ but also provides an assessment of uncertainty across the input space. 

Within Bayesian optimization, the acquisition function assumes a pivotal role, delicately balancing the interplay between exploitation and exploration. Leveraging the uncertainty estimation from the surrogate model's posterior, the acquisition function, denoted as $a_{p(g)}$, guides the selection of the next sampling point, which potentially enhances the current best-known solution. Note that the acquisition function includes Expected Improvement (EI), Probability of Improvement (PI), Upper Confidence Bound (UCB), among others. The determination of the next promising point is expressed as: $\textbf{x}_{n+1} = \mathrm{arg max} \hspace{2pt} {a}(\textbf{x})$.

% Subsequently, the recently acquired sample is assessed by the surrogate model and is re-fitted, incorporating the new observation. This process updates the posterior of the surrogate model, refining its predictive capabilities.

The final model weights are optimized by minimizing the combined forecasting error of the 3 base models, shown as:

\begin{equation*}\label{eq:bo}
\begin{aligned}
  % \begin{split}
\textbf{w}^* &= \mathrm{arg min}_{[w_1, w_2, w_3]\in \mathbb{R}^d} {g(\textbf{w})}\\
    &= \mathrm{arg min}_{[w_1, w_2, w_3]\in \mathbb{R}^d} 
    (
        \text{APE}(
            (w_1\times f_{\text{LGBM}}+ \\
            & \quad w_2\times f_{\text{PROPHET}}+w_3\times f_{\text{MSTL}}
            ), 
            y_{\text{true}} 
        )
    )
% \end{split}
\end{aligned}
\end{equation*}
$$s.t.,  w_1 + w_2 + w3= 1$$

where $w_i$ is the weight assigned to model $i$, and $\text{APE}(\cdot)$ is the average APE of the combined (model weighted) forecast. $f_{model}$ is the respective model's forecasts and $y_{\text{true}}$ is the ground truth of the time series.

The BO ensemble provides relatively robust forecasts for each level. 
However, since they are independent of the parent-children relationship, the incoherence exists within the hierarchy (children's forecasts don't add up to the parent's forecasts). 
% Though traditional reconciliation approaches like Top-Down, Bottom-Up, and Middle-Out forecasting approaches \cite{abolghasemi2022model, babai2022demand, li2018greedy, rostami2015non, zotteri2007model} can be used to make the tree coherent, they don't handle time series where there exists high variability, missing data, multi-seasonality, and randomness. 
% In our hierarchy, the first level represents the aggregated revenues from all time series at daily basis. The second level divides the total into 6 groups, according to the traffic volume and business sect of the time series. The third level further dis-aggregates the second level into the business function groups to which each fourth level time series belongs. The fourth level holds all of the most granular time series. 
% Each node in the hierarchy represent 1 time series including both its historical daily data and the future forecasts that need to be made over a time horizon.

To overcome the limitations of reconciliation such as multi-seasonality and randomness described earlier in Sec \ref{litReview}, we propose the Multi-Stage HiFoReAd framework consisting of 4 modules (Fig. \ref{fig:multistage_desc}): 
(a) Top-Down Reconciliation; 
(b) Hierarchical Harmonic Alignment Forecasts Adjustment (HHAFA); 
(c) MinTrace Reconciliation with Weighted Linear Square (WLS); 
(d) Stratified Scale-Weighted Forecasts Synchronization (SSW-FS).

First, we apply Top-Down reconciliation on the incoherent BO ensemble forecasts to ensure coherence, without altering the highest level forecasts. However, while the forecasts from the Top-Down approach are coherent, they lack accuracy and ignore the seasonality of individual children. 
To enhance seasonality and accuracy, we propose using harmonic alignment between BO ensemble forecasts and Top-Down forecasts. This improved forecast is incoherent, which is rectified by applying MinTrace reconciliation to the top three levels, ignoring the bottom level time series' noise and sparsity. 
This step enhances accuracy, maintains seasonality and ensures coherence for the top three levels. 
For the bottom level, we utilize Stratified Scale-Weighted Forecasts Synchronization to improve coherence and accuracy. Details of each step are provided below. 

\begin{figure}[h]
  \centering
  \includegraphics[height=12cm]{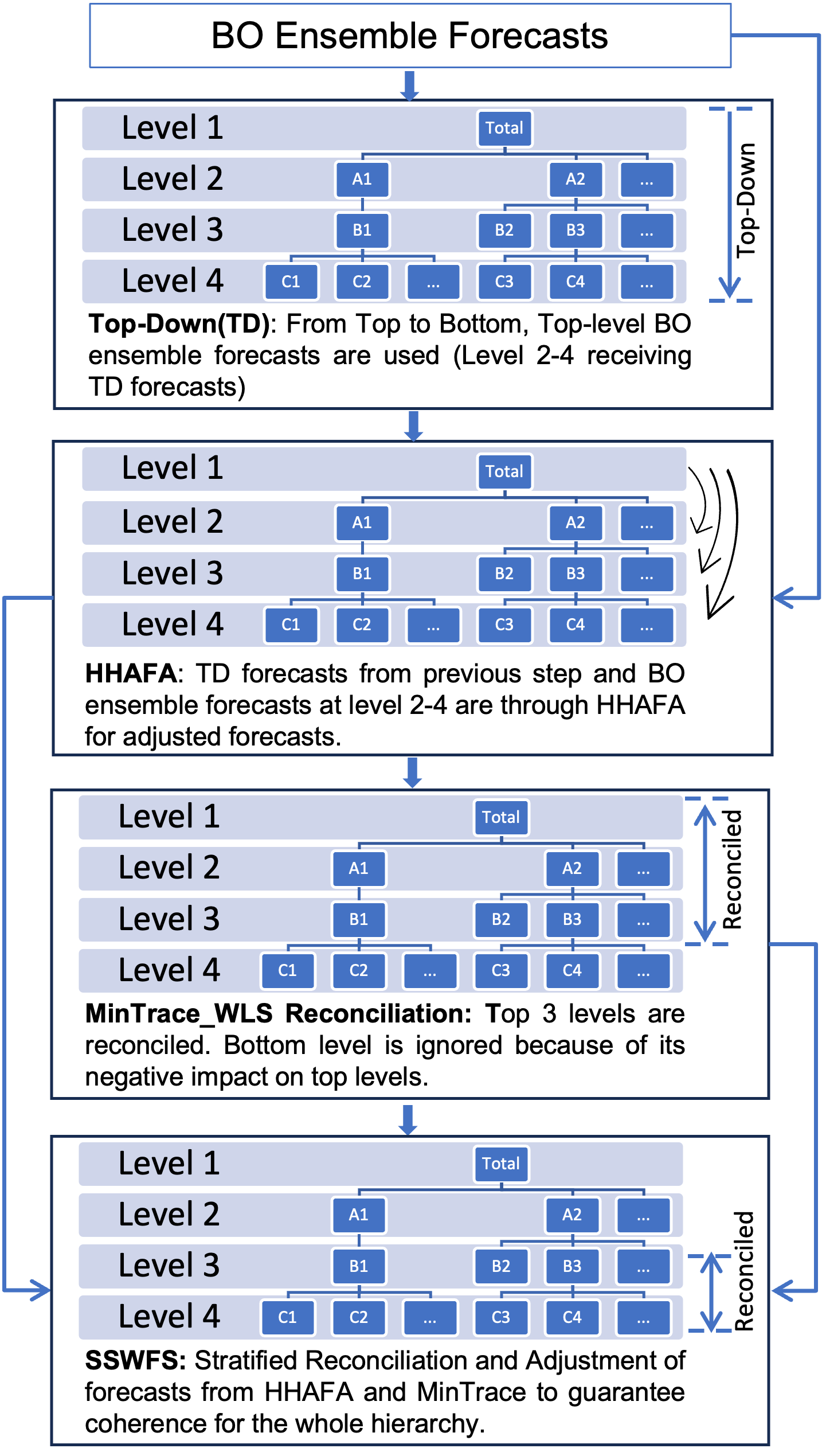}
  \caption{The Multi-Stage Hierarchical Forecasting Reconciliation and Adjustment}
  \label{fig:multistage_desc}%
\end{figure}

\textbf{(a) Top-Down Reconciliation} Ensuring coherence across levels is not guaranteed when employing Bayesian Optimization-based ensemble to generate independent time series forecasts at each level. Additionally, the data sparsity and granularity, less discernible trend and patterns, as well as increased variability issues, often associated with lower-levels (specifically the bottom level), make forecasts not as accurate as higher levels \cite{babai2022demand, mirvcetic2017modified, mircetic2022forecasting}. Therefore, in this module, we propose to utilize Top-Down approach to dis-aggregate the top level forecasts to produce the lower levels' forecasts, which are passed to the Harmonic Alignment Forecasts Adjustment module. We denote the Top-Down Reconciliation approach based on average historical proportions as follow:
\begin{equation*}
\begin{split}
\tilde{\textbf{y}}^{k}_{1i\_{TD}}(T+1:T+h|0:T) &= \\
&p^k_i * \hat{\textbf{y}}^{1}(T+1:T+h|0:T)
\end{split}
\end{equation*}
$$p^k_i=\frac{1}{L}\sum^{T}_{t=T-L+1}{\frac{y^{k}_{i}(t)}{y^{1}(t)}} \text{ where, } k \in {2,...,K-1}$$

Here, $\textbf{y}^{1}$ is the time series at the top level. Each proportion $p^k_i$ at level $k$ reflects the average of the historical proportions of the lower-level series $y^{k}_{i}(t)$ over the period $t=T-L+1,..., T$ relative to the top aggregate $\hat{y}^{1}(t)$.

\textbf{(b) Hierarchical Harmonic Alignment Forecasts Adjustment (HHAFA)} We extract the seasonality values for each time series in the hierarchy based on Fast Fourier Transform (FFT), denoted as: 
$$\textbf{s}_i = \sum_{\text{domminant frequencies}}|\vartheta_f|\cdot cos(arg(\vartheta_f))$$
$$\text{where,}\; \vartheta_f=\sum^{T-1}_{t=0}y_i(t)\cdot e^{-\frac{2\pi}{T}ft}$$ 
We remove any seasonality values larger than threshold $\tau$, as $\textbf{s}_i=\{s|s<=\tau\}$. FFT enables a rapid decomposition of time series data into frequency components to identify recurring seasonal patterns \cite{davis2016introduction, musbah2019identifying}. 

Next, we employ Jaccard similarity to assess the similarity of seasonal components between time series, $J(\textbf{s}_1, \textbf{s}_2)=\frac{|\textbf{s}_1 \cap \textbf{s}_2|}{|\textbf{s}_1 \cup \textbf{s}_2|}$. 
In order to reduce the sensitivity to minor variations in seasonal patterns and create more stable representation of the seasonality, we apply a ceiling operation $\lceil \rceil*{(c)}$ to each $\textbf{s}_i$. This helps enhance the robustness and provides consistent assessment in detecting the similarity between different time series. 
We express the HHAFA based low-level forecasts as:
\begin{equation*}
\begin{split}
&\hat{\textbf{y}}^{k}_{i}(T+1:T + h) = \\
&J(\lceil \textbf{s}^{k}_{i} \rceil, \lceil \textbf{s}^{1} \rceil) 
\cdot \tilde{\textbf{y}}^{k}_{1i\_{TD}}(T+1:T+h|0:T) \;+ \\
&(1 - J(\lceil \textbf{s}^{k}_{i} \rceil, \lceil \textbf{s}^{1} \rceil)) \cdot  \hat{\textbf{y}}^{k}_{i\_BO}(T+1:T+h|0:T), \\
\end{split}
\end{equation*}
$$k \in {2,...,K-1}$$
where, $ \hat{\textbf{y}}^{k}_{i\_BO}(T+1:T+h|0:T)$ is the BO-ensemble Forecasts while the $\tilde{\textbf{y}}^{k}_{1i\_{TD}}(T+1:T+h|0:T)$ represents the Top-Down induced Low level forecasts. HHAFA helps capture similar seasonal patterns among time series (limiting the impact of irrelevant fluctuations), allows low-level forecasts to benefit from their relationship with parental forecasts. 

\textbf{(c) Minimum Trace Reconciliation with Weighted Linear Square (WLS)}
To illustrate Minimum Trace (MinTrace) Reconciliation \cite{wickramasuriya2018optimal}, an example hierarchy is used, as shown in Fig. \ref{fig:hf_desc}. 
In Fig. \ref{fig:hf_desc}(a), there are three levels in the hierarchy with a total of 8 time series - 1 in Levels 1, 2 in level 2, and 5 in level 3.
The forecasts produced for each level's time series are known as \textit{base forecasts}. 
Fig. \ref{fig:hf_desc}(b) illustrates the matrix equation of the reconciliation for hierarchical time series, where $\textbf{S}$ is the aggregation matrix used to aggregate the bottom-level (in this example, level 3). 
Let $\hat{\textbf{b}}_t$ denote the vector of bottom-level forecasts at time $t$, where $\hat{\textbf{b}}_t = [\hat{y}_{XA,t}, \hat{y}_{XB,t}, \hat{y}_{XC,t}, \hat{y}_{YA,t}, \hat{y}_{YB,t}]$. 

The historical time series are coherent across the hierarchy, thus $\textbf{y}_t =\textbf{S} \textbf{b}_t$. 
However, the base forecasts are not coherent, $\hat{\textbf{y}}_t \neq \textbf{S} \hat{\textbf{b}}_t$. 
The reconciliation methods seek to use $\tilde{\textbf{b}}_t = \textbf{P}\hat{\textbf{y}}_t$ to receive $\textbf{bottom-level  reconciled  forecasts}$, where $\textbf{P}$ is a transformation matrix that maps the \textit{base forecasts} to the $\textbf{bottom-level  reconciled}$ $\textbf{forecasts}$. 
Then, the coherent forecasts across the whole hierarchy can be expressed as $\tilde{\textbf{y}}_t = \textbf{SP}\hat{\textbf{y}}_t = \textbf{S}\tilde{\textbf{b}}_t$, meaning aggregating the $\textbf{bottom-level  reconciled  forecasts}$ to received $\textbf{all-level}$ $\textbf{reconciled  forecasts}$. 
Here, $\hat{\textbf{y}}_t$ and $\tilde{\textbf{y}}_t$ represent the \textit{base forecasts} and the \textit{reconciled  forecasts}, respectively. 
The transformation matrix $\textbf{P}$ can be calculated by various algorithms. In our paper, we adopt MinTrace with WLS, achieved by:
$$\tilde{\textbf{Y}}(T+1:T+h)=\textbf{SP}\hat{\textbf{Y}}(T+1:T+h)$$
$$\text{given,} \; \textbf{P}=(\textbf{S}^T\textbf{W}^{-1}_h\textbf{S})^T\textbf{S}^T\textbf{W}^{-1}_h$$
$\textbf{W}_h$ is the variance-covariance matrix of the \textit{base forecasts} errors. There are various methods to estimate $\textbf{W}_h$. In our study, we adopt WLS since it gives the best results.

After HHAFA stage, the forecasts across the hierarchy are not coherent. In this stage we make the top three levels' (1, 2 and 3) forecasts coherent based on MinTrace with WLS, skipping the level 4 forecasts. 
By excluding the bottom level forecasts we mitigate its negative impact in the reconciliation process. Reconciling the more stable and aggregated higher level forecasts improve the overall accuracy of our framework.

\begin{figure}[h]
  \centering
  \includegraphics[width=\linewidth]{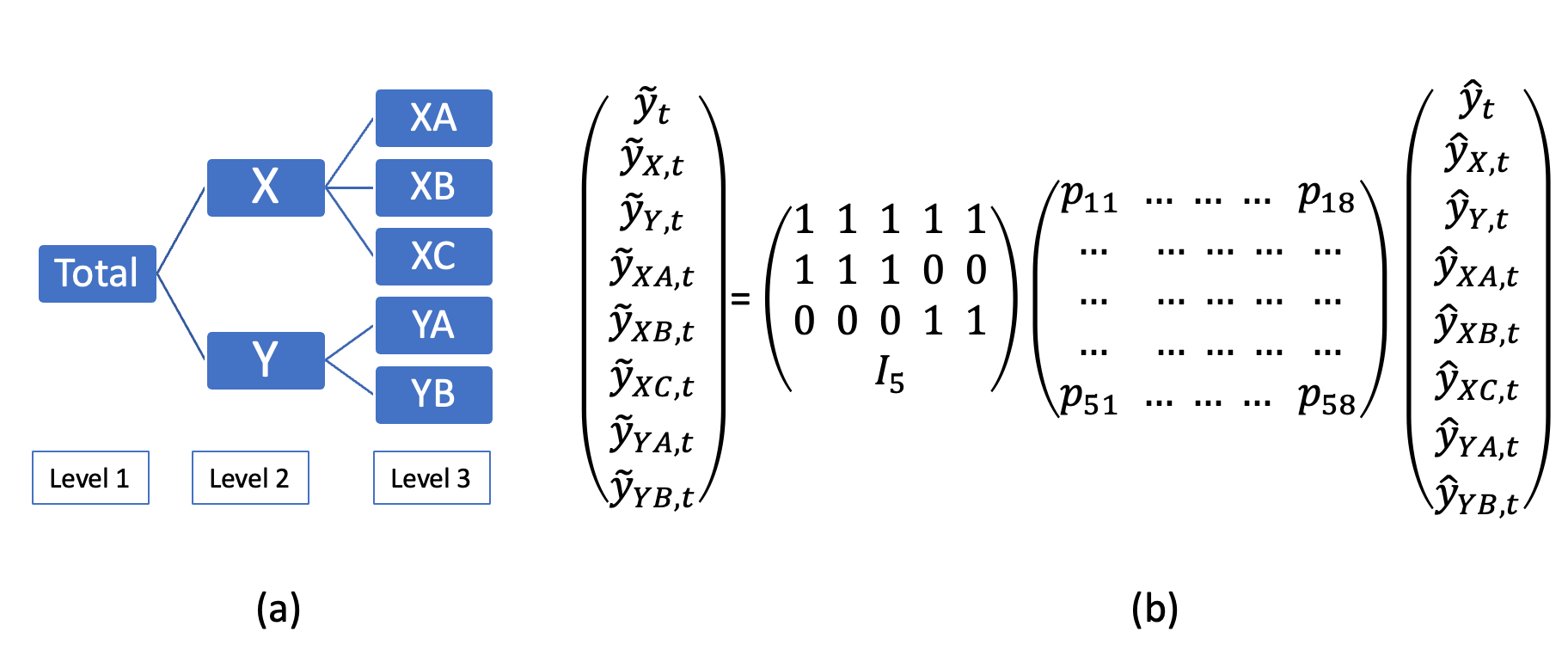}
  \caption{The Hierarchical Time Series Structure in a matrix formulation. (a) A simple hierarchy with three levels and 8 time series. (b) Matrix formulation from base forecasts to reconciled forecasts, $\hat{Y}_t = SP\Tilde{Y}_t$, where $P$ maps the base forecasts ($\Tilde{Y}_t$) to bottom-level forecasts and aggregation constrains $S$ sumps these forecasts to a set of coherent forecasts ($\hat{Y}_t$). }
   \label{fig:hf_desc}%
\end{figure}

\textbf{(d) Stratified Scale-Weighted Forecasts Synchronization (SSW-FS)} The top three levels' forecasts are coherent after stage (c). In order to improve the forecasting accuracy and make the forecasts coherent for level 4, which have the most noisy data, we introduce SSW-FS component. The SSW-FS module constitutes the final component within our proposed framework. 
This module operates specifically on the forecasts for the bottom 2 levels (level 3 and 4).

We treat each sub-hierarchical structure independently, as depicted in Fig. \ref{fig:SSW-FS_fig}. Within each 2-level sub-hierarchy, we apply HHAFA to generate the reconciled bottom-level forecasts as: 
\begin{equation*}
\begin{split}
&\hat{\textbf{y}}^{K}_{j}(T+1:T+h) = \\
&J(\lceil \textbf{s}^{K-1}_{i} \rceil, \lceil \textbf{s}^{K}_{j} \rceil) \cdot \tilde{\textbf{y}}^{K-1}_{ij\_{TD}}(T+1:T+h) + \\
&(1 - J(\lceil \textbf{s}^{K-1}_{i} \rceil, \lceil \textbf{s}^{K}_{j} \rceil)) \cdot  \hat{\textbf{y}}^{K}_{j\_BO}(T+1:T+h), j \in Childen(i)
\end{split}
\end{equation*}

\begin{figure}[h]
  \centering
  \includegraphics[width=\linewidth]{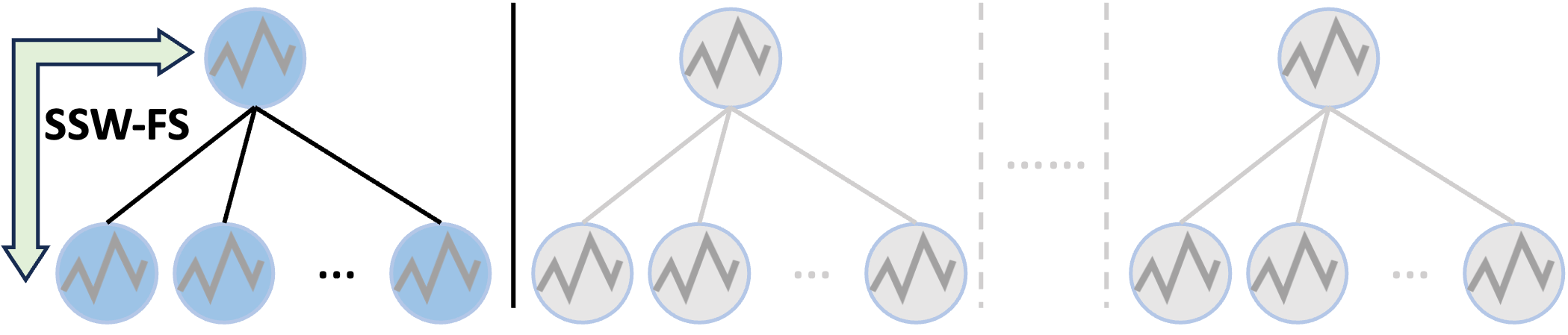}
  \caption{Stratified Scale-Weighted Forecasts Synchronization}
  \label{fig:SSW-FS_fig}%
\end{figure}

The resulting aggregated bottom-level forecasts are:
$$\hat{\textbf{y}}^{agg}(T+1:T+h)=\sum^J_{j=1}\hat{\textbf{y}}^{K}_{j}(T+1:T+h)$$
Subsequently, a scaling operation is applied to each bottom level forecast as:
\begin{equation*}
\begin{split}
&\tilde{\textbf{y}}^{SSW-FS}_{j}(T+1:T+h) =\\ 
&\tilde{\textbf{y}}_{i}(T+1:T+h) \cdot \frac{\hat{\textbf{y}}^{K}_{j}(T+1:T+h)}{\hat{\textbf{y}}^{agg}(T+1:T+h)}
\end{split}
\end{equation*}
This step ensures that the scaled forecasts maintain the relative weights of each bottom-level forecast and consequently coherence. We describe our proposed method in Algorithm \ref{algo:main_algo}.

\section{Data Description}
In this section we describe the datasets used for validating our findings. We use one real world in-house private dataset. Additionally, we use three public datasets to corroborate our framework on different domains and forecasting horizons. These four datasets covers a wide range of frequencies and are summarised in Table \ref{tab:data_table} along with their descriptive statistics.

\label{sectionDataDesc}

\begin{table*}[phtb]
  \centering
  \caption{Description of Data}
    \begin{tabular}{llcccc}
    \hline
        \multicolumn{2}{l}{Level} & \multicolumn{1}{l}{\begin{tabular}{@{}l}Number of \\ Time series\end{tabular}} &  \multicolumn{1}{l}{\begin{tabular}{@{}l}Number of non-stationary \\ Time series\end{tabular}} & \multicolumn{1}{l}{\begin{tabular}{@{}l}Percentage of \\missing values\end{tabular}} & \multicolumn{1}{l}{\begin{tabular}{@{}l}Percentage of time series inactive \\ in the last validation period \end{tabular}}\\ 
        
        \midrule
        \multicolumn{6}{c}{Walmart Connect Ads Demand Data} \\
        \midrule
        \multicolumn{2}{l}{Level 1 - Total} & 1  & 1   & 0\% &  0\% \\
        \multicolumn{2}{l}{Level 2 - Business Sect} & ${\sim}10$   & ${\sim}10$     & 0\%	& 0\%\\
        \multicolumn{2}{l}{Level 3 - Product Type} & ${\sim}30$  & ${\sim}20$  & 25\%	& 0\%  \\
        \multicolumn{2}{l}{Level 4 - Product} & ${\sim}12.8K$ & ${\sim}7.8K$ & 62\%	& 6\%  \\

        \midrule
        \multicolumn{6}{c}{Public Datasets} \\
        
        % \midrule
        % \multirow{4}{*}{M5} & Level 1 & 1 & ? & ?\% & ?\% \\ 
        %  & Level 2 & 7 & ? & ?\% & ?\% \\ 
        %  & Level 3 & 69 & ? & ?\% & ?\% \\ 
        %  & Level 4 & 15404 & ? & ?\% & ?\% \\ 
        % \hline
        
        % \multirow{4}{*}{Road Traffic} & Level 1 & 1 & ? & ?\% & ?\% \\ 
        %  & Level 2 & 2 & ? & ?\% & ?\% \\ 
        %  & Level 3 & 4 & ? & ?\% & ?\% \\ 
        %  & Level 4 & 963 & ? & ?\% & ?\% \\ 
        % \hline
        
        % \multirow{4}{*}{Tourism} & Level 1 & 1 & ? & ?\% & ?\% \\ 
        %  & Level 2 & 4 & ? & ?\% & ?\% \\ 
        %  & Level 3 & 32 & ? & ?\% & ?\% \\ 
        %  & Level 4 & 304 & ? & ?\% & ?\% \\ 

        ~ & ~ & \begin{tabular}{p{.6cm}p{.5cm}p{.5cm}}
            D1 & D2 & D3  \\
        \end{tabular}
        & \begin{tabular}{p{.6cm}p{.5cm}p{.5cm}}
            D1 & D2 & D3  \\
        \end{tabular}  
       & \begin{tabular}{p{.6cm}p{.5cm}p{.5cm}}
            D1 & D2 & D3  \\
        \end{tabular} 
        & \begin{tabular}{p{.6cm}p{.5cm}p{.5cm}}
            D1 & D2 & D3  \\
        \end{tabular} 
        \\ \hline
        \multicolumn{2}{l}{Level 1} 
        & \begin{tabular}{p{.6cm}p{.5cm}p{.5cm}}
            1 & 1 & 1  \\
        \end{tabular}
        & \begin{tabular}{p{.6cm}p{.5cm}p{.5cm}}
            1 & 0 & 1  \\
        \end{tabular}  
        & \begin{tabular}{p{.6cm}p{.5cm}p{.5cm}}
            0\% & 0\% & 0\%  \\
        \end{tabular} 
        & \begin{tabular}{p{.6cm}p{.5cm}p{.5cm}}
            0\% & 0\% & 0\%  \\
        \end{tabular} 
        \\
        \multicolumn{2}{l}{Level 2} 
        & \begin{tabular}{p{.6cm}p{.5cm}p{.5cm}}
            7 & 2 & 4  \\
        \end{tabular}
        & \begin{tabular}{p{.6cm}p{.5cm}p{.5cm}}
            5 & 0 & 4  \\
        \end{tabular}  
        & \begin{tabular}{p{.6cm}p{.5cm}p{.5cm}}
            0.1\% & 0\% & 0\%  \\
        \end{tabular} 
        & \begin{tabular}{p{.6cm}p{.5cm}p{.5cm}}
            0\% & 0\% & 0\%  \\
        \end{tabular} 
        \\
        \multicolumn{2}{l}{Level 3} 
        & \begin{tabular}{p{.6cm}p{.5cm}p{.5cm}}
            69 & 4 & 32  \\
        \end{tabular}
        & \begin{tabular}{p{.6cm}p{.5cm}p{.5cm}}
            40 & 0 & 28  \\
        \end{tabular}  
        & \begin{tabular}{p{.6cm}p{.5cm}p{.5cm}}
            2\% & 0\% & 0\%  \\
        \end{tabular} 
        & \begin{tabular}{p{.6cm}p{.5cm}p{.5cm}}
            0\% & 0\% & 0\%  \\
        \end{tabular} 
        \\
        \multicolumn{2}{l}{Level 4} 
        & \begin{tabular}{p{.6cm}p{.5cm}p{.5cm}}
            15404 & 963 & 304  \\
        \end{tabular}
        & \begin{tabular}{p{.6cm}p{.5cm}p{.5cm}}
            3856 & 0 & 144  \\
        \end{tabular}  
        & \begin{tabular}{p{.6cm}p{.5cm}p{.5cm}}
            50\% & 0.03\% & 9\%  \\
        \end{tabular} 
        & \begin{tabular}{p{.6cm}p{.5cm}p{.5cm}}
            0.6\% & 0\% & ${\sim}0\%$  \\
        \end{tabular} 
        \\
        \midrule
        \multicolumn{6}{l}{D1 = M5; D2 = Road Traffic; D3 = Tourism} \\
    \bottomrule
    \end{tabular}%
  \label{tab:data_table}%
\end{table*}%

\begin{algorithm}
\small
    \caption{Multi-Stage Hierarchical Forecasting Reconciliation and Adjustment}
  \begin{algorithmic}[1]
    \REQUIRE Reconciled forecasts for time series at each level in the hierarchy
    
    \INPUT Trained Models producing $h$-step forecasts and trained on $T$-steps,
            train, validation and test data with $K=4$ level of hierarchy, ${N_{L_k}}$ representing the number of time series at the $k$th level, \textbf{M} = [\textbf{PROPHET, LGBM, MSTL}]: 3 base models
    
    \OUTPUT Final reconciled forecasts for time series for each level $\Tilde{\textbf{Y}}^k(T+1:T+h)$.
    \STATE \textbf{Initialization} $\Tilde{\textbf{y}}^k_i(T+1:T+h) = \{\}$
    \FOR{$k = 1 \hspace{4pt} \textbf{to} \hspace{4pt} K$}
        \FOR{$i = 1 \hspace{4pt} \textbf{to} \hspace{4pt} {N_{L_k}}$}
        \STATE $\hat{\textbf{y}}_{i}^k(T+1:T+h) = []$
        \FOR{ $m = 1 \hspace{4pt} \textbf{to} \hspace{4pt} M$}
            \STATE $\hat{\textbf{y}}_{i\_m}^k(T+1:T+h)  \leftarrow  \textbf{MODEL}_m(\textbf{y}_{i\_model}^k(0:T))$\\
             \COMMENT{$h$ step forecasts from the three base models}
             \STATE $\hat{\textbf{y}}^k_{i}(T+1:T+h). \textbf{ADD}(\hat{\textbf{y}}_{i\_m}^k(T+1:T+h) )$
        \ENDFOR

  \STATE $\hat{\textbf{y}}^{k}_{i}(T+1:T+h)\leftarrow$ \textbf{BO\_ENSEMBLE}($\hat{\textbf{y}}_{i}^k(T+1:T+h)$) \\
  \COMMENT{Weighted ensemble of the M models' forecasts from optimal weights based on BO}
\ENDFOR
\ENDFOR\\
\COMMENT {$\hat{\textbf{Y}}^{k}(T+1:T+h) = [\hat{\textbf{y}}^{k}_i(T+1:T+h),... ,\hat{\textbf{y}}^{k}_{N_{L_k}}(T+1:T+h)]$}
      \FOR {$k  = 1 \hspace{4pt} \textbf{to} \hspace{4pt} K$}
        \STATE $\Tilde{\textbf{Y}}^{k}_{{TD}}(T+1:T+h) = \textbf{TOPDOWN\_REC}(\hat{\textbf{Y}}^{k}(T+1:T+h)) $\\
        \COMMENT{BO_ensemble forecasts are reconciled by Top-Down Approach}
        \STATE $\hat{\textbf{Y}}^{k}_{{HHAFA}}(T+1:T+h) = \textbf{HHAFA}(\Tilde{\textbf{Y}}^{k}_{{TD}}(T+1:T+h))$\\
        \COMMENT{Forecasts accuracy are improved by HHAFA approach}
    \ENDFOR
    \FOR{$k  = 1 \hspace{4pt} \textbf{to} \hspace{4pt} (K-1)$}
        \STATE $\Tilde{\textbf{Y}}^{k}_{{MinT}}(T+1:T+h) = \textbf{MINTRACE}(\hat{\textbf{Y}}^{k}_{{HHAFA}}(T+1:T+h)) $\\
        \COMMENT{Top three levels' forecasts are coherent and improved by using Min_Trace WLS hierarchical approach}
    \ENDFOR
    \STATE $\Tilde{\textbf{Y}}^{K}_{{SSW\text{-}FS}}(T+1:T+h) = \textbf{SSW\text{-}FS}(\hat{\textbf{Y}}^{K}_{{HHAFA}}(T+1:T+h)) $\\
    \COMMENT{Bottom level forecasts are also coherent with accuracy improved by SSW-FS approach}
    \FOR{$k = 1 \hspace{4pt} \textbf{to} \hspace{4pt} (K - 1)$}
      \STATE $\Tilde{\textbf{Y}}^k(T+1:T+h) =\Tilde{\textbf{Y}}^{k}_{{MinT}}(T+1:T+h) $
      \ENDFOR

      \STATE$\Tilde{\textbf{Y}}^K(T+1:T+h) = \Tilde{\textbf{Y}}^{K}_{{SSW-FS}}(T+1:T+h)$
  \end{algorithmic}
  \label{algo:main_algo}
\end{algorithm}

% \subsection{Walmart Connect Ads Demand Data}
\textbf{A. Walmart Connect Ads Demand Data} 
This internal dataset represents the advertising business in the growing e-commerce sector, and comprises four levels. At the top level, the time series represents the aggregated sum of demands for all products. The Business Sect at level two encompasses about 10 time series, followed by the Product Type level with 30, and the Product level with a substantial 12.8 thousand time series, with 1,400 daily observations. Table \ref{tab:data_table} shows the count of non-stationary time series, percentage of missing values at each level, and percentage of time series at each level that do not have any data in the most recent month in the dataset, which indicates that this data is particularly sparse and/or noisy. 

% \subsection{Road Traffic Data}
\textbf{B. Road Traffic Data}
This public dataset\footnote{archive.ics.uci.edu/ml/datasets/PEMS-SF} gives the occupancy rate (between 0 and 1) of 963 car lanes of San Francisco bay area freeways. This data, notably used in classification tasks \cite{cuturi2011fast}, contains measurements from 963 sensors sampled every 10 minutes from 2008-01-01 to 2009-03-30 (440 days). We aggregate the data to a daily level obtaining 963 bottom level time series of length 440. We then constructed a 4-level hierarchy by aggregating them into four groups randomly, resulting in 4 series. These 4 series were then aggregated in groups of two to obtain 2 series and finally the top level series.

% \subsection{M5}
\textbf{C. M5}
This dataset, is a comprehensive public time-series dataset focused on forecasting Walmart product sales. The dataset was provided by Walmart and hosted by Kaggle\footnote{kaggle.com/competitions/m5-forecasting-accuracy}. It contains daily sales data for around 3,000 products across 10 off-line stores in the United States over 5 years, ranging from 2011-01-29 to 2016-06-19. In our experiment, we use the sales data, ignoring the price information. In the sales data, each item is associated with an item id, a department id, a category id to which each item belongs, a store id, a state id, as well as calendar ids. We match the calendar ids with calendar table to create unit sales time series of each item. In addition, in order to align with the design and structure of the framework on our internal Walmart Connect Ads Demand Dataset, we ignore the categories ``state\_id'' and ``category\_id'' while keeping only ``department\_id'', ``store\_id'' and ``item\_id'', to build a similar 4-level structure hierarchy of time series. 

% \subsection{Tourism}
\textbf{D. Tourism}
The last dataset in our experiment is Tourism\footnote{kaggle.com/datasets/luisblanche/quarterly-tourism-in-australia}, which contains quarterly visitor data in different regions of Australia. The dataset is structured in a hierarchical format having levels such as tourist types, states, regional, etc. We aggregate all tourist counts for Australia at the top level. The second level breaks down the aggregate by the tourist types, such as visiting, business etc. The third level is related to the state or territory, such as New South Wales, Queensland etc. At the fourth level, the third level data is further divided into specific regions within each state. Instead of forecasting daily values, in this hierarchical dataset, we focus on forecasting quarterly values with the forecasting horizon being 13, on this dataset containing 80 values for each time series. 

% \hfill\\
Stationarity of time series is assessed using the Augmented Dickey Fuller test (ADF) test. Table \ref{tab:data_table} highlights that a majority at each level are non-stationary, necessitating intricate model tuning to capture both trend and non-stationary elements. For instance, Walmart's Ads data at levels 3 ($25\%$) and 4 ($62\%$) have multiple time series with missing values, making forecasting at this level especially challenging.

\section{Experiments}
\label{sectionExperiment}

In this section, we delve into the performance analysis, focusing on different components/stages through the proposed framework and the comparison between the proposed framework and SOTA models as well as reconciliation methods. 

\subsection{Setup and Metrics}
Our experiments involve the ensemble forecasts from LGBM, MSTL, and Prophet models, with the ensemble weights optimized based on Bayesian Optimization. This BO-ensemble approach seeks to leverage the unique strength from each model while mitigating their respective limitations. The optimization process yields optimized weights for these finely-tuned \textit{base models}. By leveraging the hierarchy, we aim to exploit the hierarchical dependencies to improve the overall forecasting performance. We use 4 cross-validation ``test'' sets to measure the mean and median of the APEs at each level for the aforementioned components/stages in our framework. The Multi-Stage HiFoReAd process is crucial in guaranteeing both the accuracy and coherence for the forecasts. 

To assess the performance of the forecasts at each hierarchical level, we calculate the average APEs across 4 cross-validation ``test'' sets. This also serves as the business metric used by end users for assessment of forecast quality. 

% $$APE = \frac{1}{N_{L_k}}\sum_{i=1}^{N_{L_k}}\frac{|\sum_{t=T+1}^{T+h}\tilde{\textbf{y}}^{k}_i(t)-\sum_{t=T+1}^{T+h}\textbf{y}^{k}_i(t)|}{|\sum_{t=T+1}^{T+h}\textbf{y}^{k}_i(t)|}$$
$$ APE_i^{L_k} = \frac{1}{4}\sum_{v=1}^{4}\frac{|\sum_{t=T+1}^{T+h}\tilde{\textbf{y}}^{(v)}_{i}(t)-\sum_{t=T+1}^{T+h}\textbf{y}^{(v)}_i(t)|}{|\sum_{t=T+1}^{T+h}\textbf{y}^{(v)}_i(t)|}$$

where, $\textbf{y}^{(v)}_i(t)$ and $\tilde{\textbf{y}}^{(v)}_{i}(t)$ represents the ground truth and forecasts at time $t$ in cross-validation period $v$ for any time series $i$ . The forecasting horizons are $h = 90$ for Walmart Ads dataset, 90 for M5, 13 for Tourism and 70 for Traffic. 
The Mean and Median APE for each level $k$ are:
$$Mean_{APE}^{L_k} = \frac{1}{N_{L_k}}\sum_{i=1}^{N_{L_k}}APE_i^{L_k}$$ 
$$Median_{APE}^{L_k} = \text{Median}([APE_1^{L_k}, APE_2^{L_k},...,APE_{N_{L_k}}^{L_k}])$$

\begin{table*}[!ht]
    \centering
    \caption{APE at different stages of HiFoReAd. Lowest APE is highlighted in \textcolor[rgb]{ 1,  0,  0}{red}. Coherent forecasts are marked with $\star$. The percentage improvement in APE of final reconciled forecasts compared to un-reconciled BO ensemble is listed in (\ldots).}
    \begin{tabular}{l|l|ll|ll|ll|ll}
    \toprule
        Dataset & \multirow{2}{*}{
            % Stages of Multi-Stage HiFoReAd
            \begin{tabular}{@{}l@{}}Stages of \\ Multi-Stage HiFoReAd\end{tabular}
        } & \multicolumn{2}{c|}{Level 1} & \multicolumn{2}{c|}{Level 2} & \multicolumn{2}{c|}{Level 3} & \multicolumn{2}{c}{Level 4}\\
        ~ & ~ & Mean & Median & Mean & Median & Mean & Median & Mean & Median \\ 
        \midrule
        
        % \multirow{6}{*}{\begin{tabular}{@{}l@{}}Walmart Connect\\Sales Data\end{tabular}}
        \multirow{6}{*}{\begin{tabular}{@{}@{}l@{}}Walmart Connect\\Ads Demand Data\\($h$=90)\end{tabular}}
        & BO ensemble & 4.59\% & 4.59\% & 7.95\% & 6.86\% & 20.75\% & 14.12\% & 54.83\% & 48.91\% \\ 
        ~ & TD (all levels) & 4.59\% & 4.59\% & 8.74\% & 7.01\% & 20.45\% & 14.42\% & 56.93\% & 53.77\% \\ 
        ~ & HHAFA (Level 2-4) & - & - & \textcolor[rgb]{ 1,  0,  0}{7.62\%} & 6.73\% & 20.20\% & 13.47\% & \textcolor[rgb]{ 1,  0,  0}{47.25\%} & \textcolor[rgb]{ 1,  0,  0}{41.56\%} \\ 
        ~ & MinTrace WLS (Level 1-3) & 2.74\%$^{\star}$ & 2.74\%$^{\star}$ & 8.72\%$^{\star}$ & 6.45\%$^{\star}$ & 18.55\%$^{\star}$ & 12.52\%$^{\star}$ & - & - \\ 
        ~ & SSW-FS (Level 3-4) & \textcolor[rgb]{ 1,  0,  0}{2.74\%}$^{\star}$ & \textcolor[rgb]{ 1,  0,  0}{2.74\%}$^{\star}$ & 8.72\%$^{\star}$ & \textcolor[rgb]{ 1,  0,  0}{6.45\%}$^{\star}$ & \textcolor[rgb]{ 1,  0,  0}{18.55\%}$^{\star}$ & \textcolor[rgb]{ 1,  0,  0}{12.52\%}$^{\star}$ & 48.09\%$^{\star}$ & 49.47\%$^{\star}$ \\ 
        & ~ & (40.31\%) & (40.31\%) & (-9.69\%) & (5.98\%) & (10.60\%) & (11.33\%) & (12.29\%) & (-1.14\%) \\
        \midrule

        % \multirow{6}{*}{M5}
        \multirow{6}{*}{\begin{tabular}{@{}l@{}}M5\\($h$=90)\end{tabular}}
        & BO ensemble & 2.45\% & 2.45\% & 6.94\% & 4.68\% & 6.47\% & 5.06\% & 33.71\% & 32.17\% \\
        ~ & TD (all levels) & 2.45\% & 2.45\% & 7.32\% & 5.59\% & 9.34\% & 7.03\% & 39.04\% & 37.80\% \\
        ~ & HHAFA (Level 2-4) & - & - & 7.16\% & 4.88\% & 8.72\% & 6.52\% & \textcolor[rgb]{ 1,  0,  0}{31.10\%} & 29.49\% \\
        ~ & MinTrace WLS (Level 1-3) & 2.07\%$^{\star}$ & 2.07\%$^{\star}$ & 6.56\%$^{\star}$ & 4.57\%$^{\star}$ & 6.15\%$^{\star}$ & 4.53\%$^{\star}$ & - & - \\ 
        ~ & SSW-FS (Level 3-4) & \textcolor[rgb]{ 1,  0,  0}{2.07\%}$^{\star}$ & \textcolor[rgb]{ 1,  0,  0}{2.07\%}$^{\star}$ & \textcolor[rgb]{ 1,  0,  0}{6.56\%}$^{\star}$ & \textcolor[rgb]{ 1,  0,  0}{4.57\%}$^{\star}$ & \textcolor[rgb]{ 1,  0,  0}{6.15\%}$^{\star}$ & \textcolor[rgb]{ 1,  0,  0}{4.53\%}$^{\star}$ & 32.45\%$^{\star}$ & \textcolor[rgb]{ 1,  0,  0}{28.92\%}$^{\star}$ \\ 
        & ~ & (15.29\%) & (15.29\%) & (5.43\%) & (2.50\%) & (4.87\%) & (10.61\%) & (3.76\%) & (10.12\%) \\
        \midrule

        % \multirow{6}{*}{Road Traffic}
        \multirow{6}{*}{\begin{tabular}{@{}l@{}}Road Traffic\\($h$=70)\end{tabular}}
        & BO ensemble & 2.74\% & 2.74\% & 3.31\% & 3.31\% & 3.82\% & 4.03\% & 8.61\% & 6.58\% \\ 
        ~ & TD (all levels) & 2.74\% & 2.74\% & 3.23\% & 3.23\% & 3.79\% & 3.98\% & 8.08\% & 5.82\% \\ 
        ~ & HHAFA (Level 2-4) & - & - & 3.26\% & 3.26\% & 3.80\% & 4.00\% & \textcolor[rgb]{ 1,  0,  0}{7.03\%} & \textcolor[rgb]{ 1,  0,  0}{4.35\%} \\
        ~ & MinTrace WLS (Level 1-3) & 1.69\%$^{\star}$ & 1.69\%$^{\star}$ & 2.88\%$^{\star}$ & 2.88\%$^{\star}$ & 2.74\%$^{\star}$ & 2.64\%$^{\star}$ & - & - \\ 
        ~ & SSW-FS (Level 3-4) & \textcolor[rgb]{ 1,  0,  0}{1.69\%}$^{\star}$ & \textcolor[rgb]{ 1,  0,  0}{1.69\%}$^{\star}$ & \textcolor[rgb]{ 1,  0,  0}{2.88\%}$^{\star}$ & \textcolor[rgb]{ 1,  0,  0}{2.88\%}$^{\star}$ & \textcolor[rgb]{ 1,  0,  0}{2.74\%}$^{\star}$ & \textcolor[rgb]{ 1,  0,  0}{2.64\%}$^{\star}$ & 7.60\%$^{\star}$ & 4.95\%$^{\star}$ \\ 
        & ~ & (38.38\%) & (38.38\%) & (13.01\%) & (13.01\%) & (28.31\%) & (34.56\%) & (11.74\%) & (24.73\%) \\
        \midrule

        % \multirow{6}{*}{Tourism}
        \multirow{6}{*}{\begin{tabular}{@{}l@{}}Tourism\\($h$=13)\end{tabular}}
        & BO ensemble & 6.43\% & 6.43\% & \textcolor[rgb]{ 1,  0,  0}{5.48\%} & \textcolor[rgb]{ 1,  0,  0}{5.76\%} & 9.40\% & 9.40\% & 18.53\% & 13.52\% \\ 
        ~ & TD (all levels) & 6.43\% & 6.43\% & 5.89\% & 6.38\% & 8.79\% & 8.15\% & 19.87\% & 15.20\% \\ 
        ~ & HHAFA (Level 2-4) & - & - & 5.89\% & 6.38\% & 8.79\% & 8.15\% & 19.84\% & 15.20\% \\
        ~ & MinTrace WLS (Level 1-3) & 6.13\%$^{\star}$ & 6.13\%$^{\star}$ & 5.89\%$^{\star}$ & 6.38\%$^{\star}$ & 8.79\%$^{\star}$ & 8.15\%$^{\star}$ & - & - \\ 
        ~ & SSW-FS (Level 3-4) & \textcolor[rgb]{ 1,  0,  0}{6.13\%}$^{\star}$ & \textcolor[rgb]{ 1,  0,  0}{6.13\%}$^{\star}$ & 5.89\%$^{\star}$ & 6.38\%$^{\star}$ & \textcolor[rgb]{ 1,  0,  0}{8.79\%}$^{\star}$ & \textcolor[rgb]{ 1,  0,  0}{8.15\%}$^{\star}$ & \textcolor[rgb]{ 1,  0,  0}{17.71\%}$^{\star}$ & \textcolor[rgb]{ 1,  0,  0}{9.07\%}$^{\star}$ \\ 
        & ~ & (4.67\%) & (4.67\%) & (-7.51\%) & (-10.76\%) & (6.43\%) & (13.24\%) & (4.44\%) & (32.95\%) \\
        
        \bottomrule
    \end{tabular}
    \label{tab:ape_des}
\end{table*}

% % Table generated by Excel2LaTeX from sheet 'Sheet1'
% \begin{table*}[thpb]
%   \centering
%   \caption{Estimation of Mean Gap at different steps}
%     \begin{tabular}{p{12.085em}|rrrr}
%     \toprule
%     {} & \multicolumn{4}{c}{Mean Absolute Gap} \\
% \cmidrule{2-5}    \multicolumn{1}{c|}{} & \multicolumn{1}{c}{Level - Total} & \multicolumn{1}{c}{Level - Account Segment} & \multicolumn{1}{c}{Level - Product} & \multicolumn{1}{c}{Level - Account} \\
%     \midrule
%     BO ensemble & 19645515.1 & 17566897.1 & 21779135.8 & 77679256.6 \\
%     Top Down & 19645515.1 & 19645512.4 & 19645511.9 & 19645447.4 \\
%     HHAFA & 19645515.1 & 19116675.3 & 19921170.5 & 36858343.2 \\
%     WLS & 16238366.7 & 16238366.7 & 16238366.7 & 16238366.7 \\
%     SSW-FS & \textcolor[rgb]{ 1,  0,  0}{16238366.7} & \textcolor[rgb]{ 1,  0,  0}{16238366.7} & \textcolor[rgb]{ 1,  0,  0}{16238366.7} & \textcolor[rgb]{ 1,  0,  0}{16238366.7} \\
%     \bottomrule
%     \end{tabular}%
%   \label{tab:mean_gap}%
% \end{table*}%

As illustrated in Table \ref{tab:ape_des}, the APE always remains unchanged at the level 1 because the Stage 1 Top-Down method and HHAFA (all levels) do not alter the top-level forecasts. However, at levels 2,  3 and 4, we start observing promising improvements in both Mean and Median APE. Although the forecasts are not coherent yet, we can see how the lower level forecasts (levels 2, 3 and 4) can be positively impacted by the proposed HHAFA component.

Once the level 2 and 3 are improved, excluding the bottom level 4 due to its sparsity, noise and the forecasts' negative impact on the top levels, we implement the reconciliation method MinTrace WLS on the top 3 levels (1, 2 and 3).
This component achieves coherency and significantly improves APEs across these top three levels, especially the median. 

After forecasts for the the top 3 levels are stabilised and improved, we apply the SSW-FS component. Though this brings a relative higher Median APE for level 4 compared to the HHAFA (all levels) stage, it maintains comparable Mean/Median APEs while guaranteeing coherence throughout.

% In summary, our comprehensive framework incorporates multi-step hierarchical forecasts, ensemble methods, similarity-based weighting, and reconciliation steps to achieve accurate and coherent demand forecasts across different hierarchy levels in the time series.

% \begin{figure}
%     \centering
%     \includegraphics[width=1.0\linewidth]{level_wise_ape.png}
%     \caption{Number of time series in each bucket of the APE statistics for each level}
%     \label{fig:ape_dist}
% \end{figure}

% In our investigation, we conducted a detailed breakdown of the number of time series at level-wise APE statistics based on the final forecasts. This analysis offers valuable insights into the distribution of forecasting accuracy across different hierarchical levels. 
% , as illustrated in Figure \ref{fig:ape_dist}. 

% \ref{fig:ape_dist} provides a comprehensive view of the number of time series at each level, and it is notable that the bottom level contains a substantial volume of time series compared to other levels. 

\subsection{Findings and Discussions}
Examining forecasting accuracy across different hierarchical levels reveals notable patterns in the distribution of Absolute Percentage Error (APE) values. Thus, We also provide a detailed analysis for the level-wise APE statistics based on the final forecasts, for each of our datasets in Table \ref{tab:ape_des}.

\subsubsection{Walmart Connect Ads Demand Data} 
Although the level 1 consists of only a single time series, this level aggregates all of the Ad products' demands, which is critical for the corporate-side evaluation and direction. At this level, the framework improves over 40\% compared with BO-ensemble.
At Level 2, the Median APE is $6.45\%$, reflecting a discerning improvement of 6\% (against BO-ensemble).
Moving to Level 3, we find that 16 out of 31 time series demonstrate APE values at most equal to the Median APE of $12.5\%$, which approximates 11\% decrease in both Mean and Median APE. 
% This implies a commendable level of forecasting accuracy within this level, where a substantial portion of time series aligns closely with the median error. 
In contrast, Level 4 housing an extensive number of time series, exhibits a more diverse distribution. 
Here, the highest Median APE observed is $49.47\%$, containing around 6 thousand time series within this percentile. The Mean APE at this level is 12.3\% lower than the BO-ensemble. The slight drop in median APE (-1.14\%) sacrifices some accuracy for coherence in order to maintain the aggregation constraint. 

Our improved ad demand forecasts have reduced manual tracking efforts by a 100\% at Walmart, and is empowering smaller advertisers — who lack dedicated in-house campaign managers — to make data-driven decisions independently.

\subsubsection{Road Traffic Data, M3, Tourism}
Table \ref{tab:ape_des} also shows the stages of forecasting performance (as well as the percentage relative improvements in PE Mean and Median over pure BO-ensemble of base models) for 3 public datasets across 4 hierarchical levels. We observe Mean APE improvements, averaged across all levels, of 7\% for M5, 23\% for Road Traffic, and 2\% for Tourism datasets, against the corresponding BO-ensemble forecasts, respectively. The observation of stage-wise forecasting improvement over different public datasets also indicates that HiFoReAd, by breaking down the reconciliation task into sub-jobs where each job handles alignment between levels, discrepancy reduction from independent predictions as well as trend and seasonal information pass among levels, can lead to better forecasting outcomes. 
This consistent pattern for all 3 datasets demonstrates the flexibility and generalization ability of HiFoReAd across domains of data, while guaranteeing coherence and maintaining the effectiveness and individual trend-seasonality patterns of time-series.

% \hfill\\

Though the multi-stage reconciliation framework introduces computational overhead, it enables parallel (and distributed) computing thus balancing this cost with faster computation, while yielding superior forecasting performance.

In addition, the APE distribution underscores an intriguing pattern — as we descend to more granular levels, the performance tends to diminish. Specifically, at level 4, where data is most granular, the forecasting accuracy is notably lower. This phenomenon can be attributed to the inherent challenges such as noisy and sparse data. The observed variations in APE across hierarchical levels highlight the importance of considering the intricacies of data granularity when interpreting forecasting results. While the aggregated levels exhibit superior performance, the granular levels necessitate additional attention and potentially specialized modeling techniques to address the unique challenges posed by sparse and noisy data. These insights contribute to a nuanced understanding of forecasting at different levels within the hierarchical structure.

% \subsubsection{Running Time Improvements}
% ...

% \begin{figure}
%     \centering
%     \includegraphics[width=1.0\linewidth]{DL_comp.png}
%     \caption{Number of time series in each bucket of the APE statistics for each level}
%     \label{fig:DL_Comp}
% \end{figure}

\subsection{Comparative Analysis}

\begin{table}
    \centering
    \caption{Comparison of Multi-Stage HiFoReAd (Ours) with other baseline Hierarchical Reconciliation frameworks. Best method is highlighted in \textcolor[rgb]{ 1,  0,  0}{red}.}
    \begin{adjustbox}{width=\columnwidth}
    \begin{tabular}{l|l|ll|ll|ll}
    \toprule
        Method & \multicolumn{1}{c|}{Level 1} & \multicolumn{2}{c|}{Level 2} & \multicolumn{2}{c|}{Level 3} & \multicolumn{2}{c}{Level 4}\\
        ~ & \multirow{2}{*}{
            \begin{tabular}{@{}l@{}}Mean / \\ Median\end{tabular}
        } & Mean & Median & Mean & Median & Mean & Median \\ 
        &  &  &  &  &  &  & \\
        
        \midrule
        \multicolumn{8}{c}{Walmart Connect Ads Demand Data} \\
        \midrule
        Ours & 2.74\% & \textcolor[rgb]{ 1,  0,  0}{8.72\%} & \textcolor[rgb]{ 1,  0,  0}{6.45\%} & \textcolor[rgb]{ 1,  0,  0}{18.55\%} & \textcolor[rgb]{ 1,  0,  0}{12.52\%} & \textcolor[rgb]{ 1,  0,  0}{48.09\%} & \textcolor[rgb]{ 1,  0,  0}{49.47\%} \\ 
        % MT OLS & 11.22\% & 15.09\% & 13.80\% & 41.25\% & 22.03\% & 83.81\% & 77.41\% \\ 
        MT WLS & \textcolor[rgb]{ 1,  0,  0}{1.76\%} & 12.50\% & 6.52\% & 21.51\% & 14.78\% & 95.83\% & 58.98\% \\ 
        OC OLS & 3.10\% & 12.06\% & 6.66\% & 22.40\% & 15.03\% & 79.81\% & 68.81\% \\ 
        % OC WLS & 11.22\% & 15.09\% & 13.80\% & 41.25\% & 22.03\% & 83.81\% & 77.41\% \\ 
        TD & 4.59\% & 8.74\% & 7.01\% & 20.45\% & 14.42\% & 56.93\% & 53.77\% \\

        \midrule
        \multicolumn{8}{c}{M5} \\
        \midrule
        Ours & \textcolor[rgb]{ 1,  0,  0}{2.07\%} & \textcolor[rgb]{ 1,  0,  0}{6.56\%} & \textcolor[rgb]{ 1,  0,  0}{4.57\%} & \textcolor[rgb]{ 1,  0,  0}{6.15\%} & \textcolor[rgb]{ 1,  0,  0}{4.53\%} & \textcolor[rgb]{ 1,  0,  0}{32.45\%} & \textcolor[rgb]{ 1,  0,  0}{28.92\%} \\
        % MT OLS & 2.31\% & 16.57\% & 4.74\% & 24.76\% & 7.12\% & 38.28\% & 37.06\% \\
        MT WLS & 3.24\% & 6.60\% & 5.20\% & 8.74\% & 7.07\% & 37.49\% & 36.24\% \\
        OC OLS & 2.31\% & 16.57\% & 4.74\% & 24.76\% & 7.12\% & 38.28\% & 37.06\% \\
        % OC WLS & 3.24\% & 6.60\% & 5.20\% & 8.74\% & 7.07\% & 37.49\% & 36.24\% \\
        TD & 2.45\% & 7.32\% & 5.59\% & 9.34\% & 7.03\% & 39.04\% & 37.80\% \\
        
        \midrule
        \multicolumn{8}{c}{Road Traffic} \\
        \midrule
        Ours & \textcolor[rgb]{ 1,  0,  0}{1.69\%} & 2.88\% & 2.88\% & \textcolor[rgb]{ 1,  0,  0}{2.74\%} & \textcolor[rgb]{ 1,  0,  0}{2.64\%} & \textcolor[rgb]{ 1,  0,  0}{7.60\%} & \textcolor[rgb]{ 1,  0,  0}{4.95\%} \\
        % MT OLS & 2.54\% & 2.56\% & 2.56\% & 3.38\% & 3.01\% & 8.78\% & 6.63\% \\
        MT WLS & 1.83\% & \textcolor[rgb]{ 1,  0,  0}{1.86\%} & \textcolor[rgb]{ 1,  0,  0}{1.86\%} & 3.19\% & 3.21\% & 7.98\% & 5.56\% \\
        OC OLS & 2.54\% & 2.56\% & 2.56\% & 3.38\% & 3.01\% & 8.78\% & 6.63\% \\
        % OC WLS & 1.83\% & 1.86\% & 1.86\% & 3.19\% & 3.21\% & 7.98\% & 5.56\% \\
        TD & 2.74\% & 3.31\% & 3.31\% & 3.82\% & 4.03\% & 8.61\% & 6.58\% \\

        \midrule
        \multicolumn{8}{c}{Tourism} \\
        \midrule
        Ours & \textcolor[rgb]{ 1,  0,  0}{6.13\%} & 5.89\% & 6.38\% & \textcolor[rgb]{ 1,  0,  0}{8.79\%} & \textcolor[rgb]{ 1,  0,  0}{8.15\%} & \textcolor[rgb]{ 1,  0,  0}{17.71\%} & \textcolor[rgb]{ 1,  0,  0}{9.07\%} \\
        % MT OLS & 6.29\% & 6.13\% & 6.07\% & 10.36\% & 9.23\% & 21.42\% & 16.72\% \\
        MT WLS & 6.22\% & \textcolor[rgb]{ 1,  0,  0}{5.79\%} & \textcolor[rgb]{ 1,  0,  0}{5.79\%} & 8.95\% & 8.34\% & 19.15\% & 14.55\% \\
        OC OLS & 6.29\% & 6.13\% & 6.07\% & 10.36\% & 9.23\% & 21.42\% & 16.72\% \\
        % OC WLS & 6.22\% & 5.79\% & 5.79\% & 8.95\% & 8.34\% & 19.15\% & 14.55\% \\
        TD & 6.43\% & 5.89\% & 6.38\% & 9.79\% & 8.70\% & 19.87\% & 15.20\% \\

        \midrule
        \multicolumn{8}{c}{MT = MinTrace; OC = Optimal Combination; TD = TopDown Forecast Proportions} \\
        
        \bottomrule
    \end{tabular}
    \end{adjustbox}
    \label{tab:baseline_rec_comparison}
\end{table}

In addition to evaluating our framework against the BO-ensemble forecasts, we also conduct comparative analysis with various baseline Hierarchical Reconciliation techniques including the state-of-the-art MinTrace,
% \cite{wickramasuriya2018optimal}
and the Optimal Combination,
% \cite{hyndman2014optimally}
and Top Down 
% \cite{gross1990disaggregation} 
methods.
The results presented in Table \ref{tab:baseline_rec_comparison} show that our framework outperforms them for most levels of the hierarchy on Walmart's Ads Demand data. Further evaluation on public datasets, presented in Table \ref{tab:baseline_rec_comparison}, demonstrates our framework's competitively smaller errors metrics (APE Mean and Median) across most or all levels compared to baseline reconciliation methods. Our framework's superior performance compared to these reconciliation methods is majorly due to the preservation of seasonality and patterns of closely related parental time series' impact on children time series by the HHAFA and SSW-FS stages.

\begin{table}
    \centering
    \caption{Comparison of Multi-Stage HiFoReAd (Ours) with state-of-the-art transformer forecasting models on Walmart Connect Ads Demand Data. Best method is highlighted in \textcolor[rgb]{ 1,  0,  0}{red}. The percentage improvement in APE of our framework compared to the DL models is highlighted in \textcolor[rgb]{ 0,  0,  1}{blue}.}
    \begin{adjustbox}{width=\columnwidth}
    \begin{tabular}{l|l|ll|ll|ll}
    \toprule
        Method & \multicolumn{1}{c|}{Level 1} & \multicolumn{2}{c|}{Level 2} & \multicolumn{2}{c|}{Level 3} & \multicolumn{2}{c}{Level 4}\\
        ~ & \multirow{2}{*}{
            \begin{tabular}{@{}l@{}}Mean / \\ Median\end{tabular}
        } & Mean & Median & Mean & Median & Mean & Median \\ 
        &  &  &  &  &  &  & \\
        \midrule
        Ours & \textcolor[rgb]{ 1,  0,  0}{2.74\%} & \textcolor[rgb]{ 1,  0,  0}{8.72\%} & \textcolor[rgb]{ 1,  0,  0}{6.45\%} & \textcolor[rgb]{ 1,  0,  0}{18.55\%} & \textcolor[rgb]{ 1,  0,  0}{12.52\%} & \textcolor[rgb]{ 1,  0,  0}{48.09\%} & \textcolor[rgb]{ 1,  0,  0}{49.47\%} \\ 
        FEDformer & 5.33\% & 9.43\% & 8.33\% & 24.97\% & 16.71\% & 70.40\% & 50.08\% \\ 
        ~ & \textcolor[rgb]{ 0,  0,  1}{(48.59\%)} & \textcolor[rgb]{ 0,  0,  1}{(7.53\%)} & \textcolor[rgb]{ 0,  0,  1}{(22.57\%)} & \textcolor[rgb]{ 0,  0,  1}{(25.71\%)} & \textcolor[rgb]{ 0,  0,  1}{(25.07\%)} & \textcolor[rgb]{ 0,  0,  1}{(31.69\%)} & \textcolor[rgb]{ 0,  0,  1}{(1.22\%)} \\
        Pyraformer & 38.45\% & 43.03\% & 42.11\% & 56.93\% & 57.57\% & 91.53\% & 92.80\% \\ 
        ~ & \textcolor[rgb]{ 0,  0,  1}{(92.87\%)} & \textcolor[rgb]{ 0,  0,  1}{(79.74\%)} & \textcolor[rgb]{ 0,  0,  1}{(84.68\%)} & \textcolor[rgb]{ 0,  0,  1}{(67.42\%)} & \textcolor[rgb]{ 0,  0,  1}{(78.25\%)} & \textcolor[rgb]{ 0,  0,  1}{(47.46\%)} & \textcolor[rgb]{ 0,  0,  1}{(46.69\%)} \\
        % Dlinear & 16.06\% & 16.06\% & 17.85\% & 14.39\% & 100.75\% & 30.41\% & 75.48\% & 70.68\% \\ 
        TiDE & 11.22\% & 15.09\% & 13.80\% & 41.25\% & 22.03\% & 83.81\% & 77.41\% \\ 
        ~ & \textcolor[rgb]{ 0,  0,  1}{(75.58\%)} & \textcolor[rgb]{ 0,  0,  1}{(42.21\%)} & \textcolor[rgb]{ 0,  0,  1}{(53.26\%)} & \textcolor[rgb]{ 0,  0,  1}{(55.03\%)} & \textcolor[rgb]{ 0,  0,  1}{(43.17\%)} & \textcolor[rgb]{ 0,  0,  1}{(42.62\%)} & \textcolor[rgb]{ 0,  0,  1}{(36.09\%)} \\
        \bottomrule
    \end{tabular}
    \end{adjustbox}
    \label{tab:sota_comparison}
\end{table}

\begin{figure}[h]
  \centering
  \includegraphics[width=\linewidth]{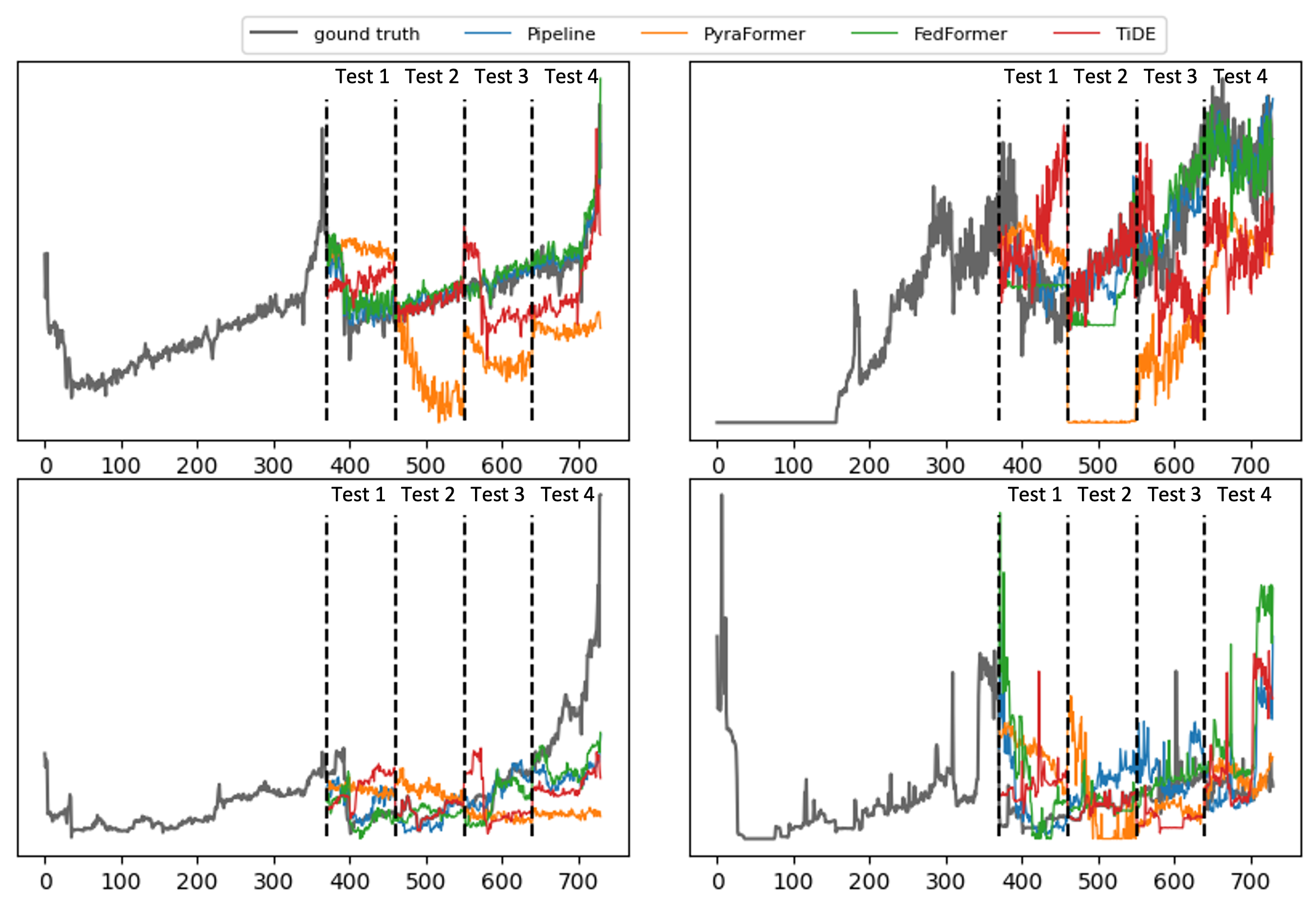}
  \caption{Sampled visualization of cross-validation forecasts between various models}
  \label{fig:sample_ts}%
\end{figure}

Further, we compare our framework against various SOTA Deep Learning models, on Walmart's Ads dataset, including Pyraformer \cite{liu2021pyraformer}, FEDformer \cite{zhou2022fedformer}, and TiDE \cite{das2023long}, each applied without the use of any Hierarchical Forecasting Reconciliation techniques. The notion that advanced models, especially NN models always outperform older ones is challenged by our experimental findings. 
The results, presented in Table \ref{tab:sota_comparison}, reveal that FEDformer exhibits promising performance among these models. 
However, it falls short of achieving the forecasting accuracy observed in our HiFoReAd framework. 
The high APE for the other models suggesting a limitation in their ability to capture the underlying seasonality and trend within our data. For example, for Level 1, the Mean APE of our method is lower than 48\%, 92.87\%, 75.58\% for FEDformer, PyraFormer and TiDE respectively, with similar trends for other levels, despite the inherent complexity and resource-intensiveness of these models. 
In Fig. \ref{fig:sample_ts} the 4 randomly sampled time series visually demonstrate the respective superior performance, in 4 cross-validation periods.  
% (In order to obfuscate sensitive business information, we remove the real y-axis values and replace dates with ordered numbers). 

Running time comparison between our proposed framework and baseline reconciliation methods on the 4 datasets indicates our framework is more scalable on large datasets such as Walmart dataset and M5, due to its ability to be distributed in Spark environment, each averaging around 0.5 and 0.4 hours running time respectively. While Top-down reconciliation method might be comparable in run-time, but MT-WLS and OC-OLS reach \~3 hours on these datasets. The running time on Traffic and Tourism datasets is less than 5 and 3 mins respectively, whereas the classic reconciliation methods take nearly 1 hr and 15 mins on the corresponding datasets. These experiments demonstrate the superior efficiency and effectiveness of our framework in handling large/complex industrial scale requests while maintaining coherent outcomes. 

Our framework is able to minimize the forecasting gaps from the ground-truth across the validation periods, and better capture the seasonality and trend in contrast to other models and reconciliation methods across all experimented datasets.

\section{Conclusion and Future Work}
The comprehensive framework, ``Multi-Stage Hierarchical Forecasting Reconciliation and Adjustment'', introduced in this paper produces coherent and seasonal forecasts with a high degree of accuracy.
By integrating advanced methodologies such as Bayesian Optimization ensemble, and multi-stage hierarchical reconciliation and adjustment modules, the framework demonstrates significant and incremental improvements in forecasting accuracy. 
Additionally, our distributed mode of modeling allows fast and efficient computation of large-scale time series data that is common in real-world applications.
Through experiments on 4 datasets, we validate the framework performance and also demonstrate the improvements in accuracy and coherence through the Multi-Stage HiFoReAd process compared to other SOTA reconciliation methods.

This enhanced end-to-end forecasting framework primarily serves as the Ads Demand Forecast at Walmart Connect, and offers practical implications for business development and AdTech teams, empowering them to make informed decisions regarding revenue planning, resource allocation, and overall business enhancement in a dynamic market environment.

Moving forward, we would explore additional use-cases of forecasting and further enhance the proposed framework, by incorporating enhanced deep learning techniques, cold-start handling and confidence interval estimations. We expect to work in hierarchical forecasting in homogeneous/heterogeneous graph and tree domains as well as to apply them in a distributed computing setting. 
% By continuously refining forecasting methodologies, organizations can better navigate market dynamics and capitalize on emerging opportunities for growth and success.

\bibliographystyle{IEEEtran}
\bibliography{bib_demand_forecasting}

\end{document}